\documentclass[sigconf]{acmart}
\usepackage{bm}

\newcommand{\x}{{\bf x}}

\newcommand{\defer}{\sc{Defer}}
\AtBeginDocument{%
  \providecommand\BibTeX{{%
    \normalfont B\kern-0.5em{\scshape i\kern-0.25em b}\kern-0.8em\TeX}}}

\setcopyright{acmcopyright}
\acmYear{2021} 
\setcopyright{acmcopyright}
\acmConference[KDD '21] {Proceedings of the 27th ACM SIGKDD Conference on Knowledge Discovery and Data Mining}{August 14--18, 2021}{Virtual Event, Singapore.}
\acmBooktitle{Proceedings of the 27th ACM SIGKDD Conference on Knowledge Discovery and Data Mining (KDD '21), August 14--18, 2021, Virtual Event, Singapore}
\acmPrice{15.00}
\acmISBN{978-1-4503-8332-5/21/08} 
\acmDOI{10.1145/3447548.3467086}

\begin{document}
\fancyhead{}
\title{Real Negatives Matter: Continuous Training with Real Negatives for Delayed Feedback Modeling}

%
\author{Siyu Gu*, Xiang-Rong Sheng*, Ying Fan, Guorui Zhou, Xiaoqiang Zhu}
\thanks{*Both authors contributed equally to this research.}
\email{{wushang.gsy,xiangrong.sxr,fanying.fy,guorui.xgr,xiaoqiang.zxq}@alibaba-inc.com}
\affiliation{%
  \institution{Alibaba Group}
  \city{Beijing, China}
}
\renewcommand{\shortauthors}{Gu and Sheng, et al.}

 \begin{abstract}
One of the difficulties of conversion rate (CVR) prediction is that the conversions can delay and take place long after the clicks. The delayed feedback poses a challenge: fresh data are beneficial to continuous training but may not have complete label information at the time they are ingested into the training pipeline. To balance model freshness and label certainty, previous methods set a short waiting window or even do not wait for the conversion signal. If conversion happens outside the waiting window, this sample will be duplicated and ingested into the training pipeline with a positive label. However, these methods have some issues. First, they assume the observed feature distribution remains the same as the actual distribution. But this assumption does not hold due to the ingestion of duplicated samples. Second, the certainty of the conversion action only comes from the positives. But the positives are scarce as conversions are sparse in commercial systems. These issues induce bias during the modeling of delayed feedback. In this paper, we propose \textbf{DE}layed \textbf{FE}edback modeling with \textbf{R}eal negatives ({\defer}) method to address these issues. The proposed method ingests real negative samples into the training pipeline. The ingestion of real negatives ensures the observed feature distribution is equivalent to the actual distribution, thus reducing the bias. The ingestion of real negatives also brings more certainty information of the conversion. To correct the distribution shift, {\defer} employs importance sampling to weigh the loss function. Experimental results on industrial datasets validate the superiority of {\defer}. {\defer} have been deployed in the display advertising system of Alibaba, obtaining over 6.0\% improvement on CVR in several scenarios. The code and data in this paper are now open-sourced\footnote{\url{https://github.com/gusuperstar/defer.git}}.
\end{abstract}





\maketitle
\section{Introduction}\label{sec:intro}
Display advertising has involved the use of the internet as a medium to obtain website traffic and deliver marketing messages~\cite{GoldfarbT2011OnlineDisplayAdvertising,Evans2009OnlineAdvertisingIndustry,HePJXLXSAHBC14PracticalLessonsFB,ZhangZWX2016BidawareGDinDisplayAdvertising,ZhangYW2014OptimalRTBforDisplayAdvertising}.
 Cost-per-conversion (CPA) is a widely used performance-dependent payment model in display advertising, where advertisers bid for predefined conversion actions, such as purchases and adding items to the cart. The systems then convert the bids to effective cost per mille (eCPM) depending on the predicted CVR. Thus, estimating accurate CVR is of importance for a competitive advertising system.
 

In display advertising, the data distribution is dynamically shifting due to the special events, change of active advertisers, ad campaigns, and users. To catch up with the distribution shift, the model often updates continuously to keep fresh~\cite{McMahan2011FTRL,KtenaTTMDHYS2019OnlineDFM}.
Most industrial systems adopt continuous training for click-through rate (CTR) and CVR prediction: the training pipeline waits for a time window before assigning a label to the sample. If a click or conversion happens within the window, it will be labeled as positive, else as negative. Samples are then ingested into the training pipeline.

However, different from CTR prediction modeling, there exists a severe \textbf{delayed feedback} problem in CVR prediction modeling: the conversion action may occur hours or days later. For samples clicked near the training timing, it might be mistakenly labeled as a negative due to the conversion waiting time being too short. Thus, although benefited from real-time training, a short waiting time window introduces \textit{fake negatives}~\cite{KtenaTTMDHYS2019OnlineDFM}, which are referred to as incorrectly labeled negative samples, while a long window keeps the correctness but loses freshness. 

\begin{figure*}[!t]
    \centering
    \includegraphics[width=\textwidth]{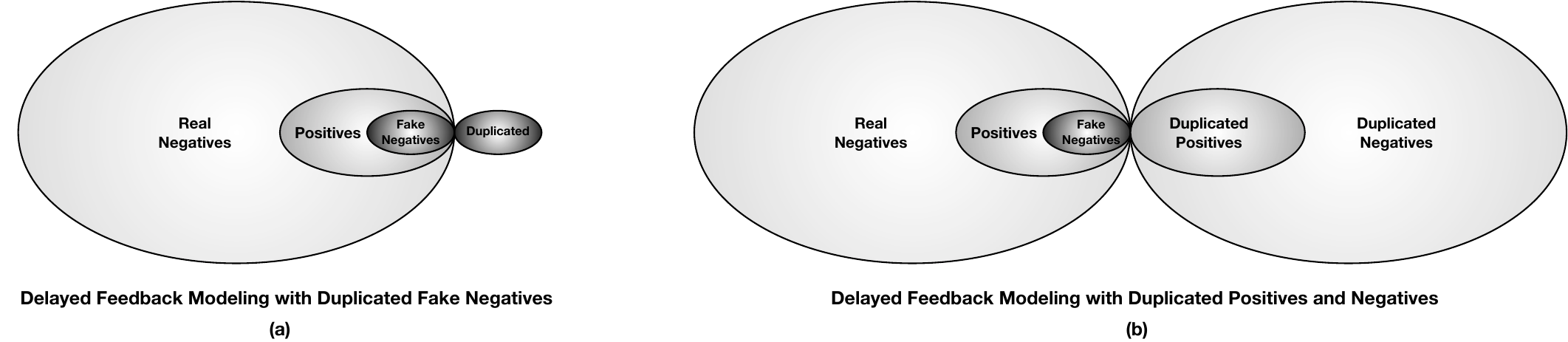}
    \caption{(a): Data pipeline of previous online delayed feedback methods, which duplicate fake negative samples with positive labels. (b) Data pipeline of the proposed method. The proposed method duplicate both positives and real negatives. After the duplication, the feature distribution remains the same as the actual distribution.}
    \label{fig:data_pipeline}
\end{figure*}

Previous methods~\cite{KtenaTTMDHYS2019OnlineDFM, YangLHZZZT2020ESDFM} address delayed feedback for continuous training by redesigning the data pipeline and loss function. To ensure the model training on the fresh data stream, they set a short window to wait for some immediate conversions or even don't wait for any delayed conversion at all. 
If the actual conversion does not occur within the waiting window, the corresponding sample will be labeled as a negative one and ingested into the training pipeline. Then once the conversion happens, the same data will be duplicated and ingested into the training pipeline with a positive label. The duplication provides more certainty of the conversion action. Since there exists a disparity between the observed distribution and the actual distribution, importance sampling methods are employed to weigh the loss function. However, previous methods still have some issues. First, the ingestion of duplicated positives changes the feature distribution. But these methods assume the observed feature distribution remains the same, which increases the bias.  Second, the certainty of the conversion action only comes from the scarce positive labels, making the model hard to learn and perform well.

In this paper, we propose \textbf{DE}layed \textbf{FE}edback modeling with \textbf{R}eal negatives ({\defer}) for online CVR prediction, which tackles the previous issues by ingesting duplicated real negatives, as shown in Figure~\ref{fig:data_pipeline}. Concretely, negatives that do not have conversions eventually will also be duplicated and ingested into the training pipeline with the negative labels. The ingestion of duplicated samples ensures that the observed feature distribution $q(\x)$ is equivalent to the actual $p(\x)$, thus reducing the bias. Moreover, the ingestion of real negatives provides more label certainty, facilitating the modeling of delayed feedback. With the ingestion of real negatives, the proposed {\defer} method employs importance sampling to correct the distribution shift. Since the delay time of real negatives equals to the maximum delay time of positive samples, the model freshness is unlikely to be affected a lot by the real negatives but bring more label certainty to the training algorithm. Experimental results on industrial datasets validate the effectiveness of the proposed {\defer} approach. We also illustrate that the ingestion of real negatives can consistently improve previous online delayed feedback methods.

Apart from the proposed method for online CVR prediction, we also share our practices on deployment for commercial platforms with different conversion \textit{attribution mechanisms} and business demands. The attribution mechanism is the mechanism used to assign credit for a conversion~\cite{WangZY2017DARTB,KtenaTTMDHYS2019OnlineDFM}. The attribution mechanism includes an attribution window: a conversion is attributed to a click only if it occurred within this time window.  
We split the deployment setting with different attribution windows and business demand into 3 types and give the corresponding solution. (1) For business scenarios with a short attribution window, e.g., 1 day, we employ continuous training with the proposed {\defer} method. (2) For business scenarios with a long attribution window, e.g., 7 days, the ingestion of duplicated samples may increase additional infrastructure cost on maintaining data caches. We set a window shorter than the attribution window to approximate the duplicated samples. We empirically found the approximation can achieve comparable performance for scenarios with a long attribution window. (3) Although continuous training is better for CVR prediction, in some industrial scenarios, the model is trained on data of a long duration daily or weekly considering the business requirements and resource consumption of real-time updates.
Here, we refer to this paradigm as offline training. Offline training loses freshness but have more correct labels. In this situation, continuous training methods are not suitable.
For commercial platforms that cannot deploy continuous training, we also give an offline training solution to address the delayed feedback problem which has been proved to be useful in some scenarios of our display advertising system. 

The main contributions of this paper are summarized as follow:
\begin{itemize}
    \item We stress the importance of real negatives in continuous CVR training. The real negatives are duplicated and ingested into the training pipeline. The ingestion of real negatives eliminates the bias in feature distribution introduced by the duplication of positive samples, which is ignored by previous methods. Moreover, it provides more certain information about the conversion signals, facilitating the modeling of delayed feedback. We further develop an importance sampling method to model the CVR.
    \item We conduct experiments on industrial datasets and deploy the algorithm in the display advertising system of Alibaba. The consistent superiority validates the efficacy of {\defer} and the ingestion of real negatives consistently improves previous delayed feedback methods. During online experiments, the deployment of the proposed method obtains over 6.0\% CVR improvement in several scenarios. The code and data for the experiments are now publicly available.
    \item We share our deployment experiences for commercial platforms with different conversion attribution mechanisms and business demands. Continuous training and offline training practices are both shared. The practical experience gained in our deployment generalizes to other setups and is thus of interest to both researchers and industrial practitioners.
\end{itemize}

\section{Related Work}
Some previous works on CVR prediction do not focus on the delayed feedback problem~\cite{MaZHWHZG2018ESMM,LeeODL2012EstimateCVR,MenonCGAK2011ResponsePred,RosalesCM2012PostCVR}. These method are not trained continuously and ignore the falsely labeled negatives. 
For example, if the model is trained on 30 day  data and the attribution window is 7 days (the delay between a click and a conversion can never be more than 7 days), then the labels of the negative samples from the last 7 days may be incorrect since the conversion may happen in the future. Note that the negative label is much more likely to be incorrect if the click has just happened. 
\subsection{Delayed Feedback Models}

In display advertising, the distribution is dynamically changing due to the change of active advertisers, ad campaigns and users. For example, when new ad campaigns are added to the system, models built on past data may not perform as well on those new ad campaigns. To capture the change in data, machine learning models should be updated continuously to keep fresh. 
To this end, during training, a waiting window can be set: the conversion is attributed to a click once it occurred within this waiting window. Then the fresh samples can be ingested by the training algorithm for continuous training. However, the continuous learning manner may introduce more falsely labeled negative samples (fake negatives) due to the short window. 
To correct the bias introduce by the fake negatives, delayed feedback models~\cite{Chapelle2014DFM,Yoshikawa2018NonparamDFM} are proposed.
The importance of delayed feedback modeling is first emphasized by DFM~\cite{Chapelle2014DFM}. In DFM, a negative is treated as unlabeled samples since the conversion has not happened yet. Besides predicting the CVR, DFM also introduces a second model that captures the expected delay between the click and the conversion and assumes the delay distribution follows the exponential distribution.  These two models are trained jointly. 
Note that delayed feedback modeling is also closely related to survival time analysis~\cite{George2003SurvivalTimeAnal} that studies the distribution of survival times.
In practice, the exponential assumption may be violated. To address this issue, \citet{Yoshikawa2018NonparamDFM} propose a non-parametric delayed feedback model for CVR prediction that represents the distribution of the time delay without assuming a parametric distribution.

\citet{KtenaTTMDHYS2019OnlineDFM} also adopt a continuous training scheme by ingesting all samples with a negative label initially. If a positive engagement occurs, this sample will be duplicated with positive label and ingested to the training algorithm for the second time. Hence, the biased observed data distribution consists of all samples from the actual data distribution labeled as negatives and the original positive samples. To learn the model from the bias distribution, \citet{KtenaTTMDHYS2019OnlineDFM} proposes two loss function FNW and FNC utilizing importance sampling~\citet{SugiyamaNKBK2007ImportanceEstimation,BottouPCCCPRSS2013Counterfactual} to handle the distribution shift. Instead of labeling all samples as negative initially, \citet{YasuiMFS2020FeedbackShift} propose a feedback shift importance weighting algorithm, in which the model waits for the real conversion in a certain time interval. However, it does not allow the data correction, i.e., duplicated positive samples even a conversion event took place in the future. To find a trade-off between the delay in model training and the fake negative rate, ESDFM~\cite{YangLHZZZT2020ESDFM}  models the relationship between the observed conversion distribution and the true conversion distribution.

\subsection{Delayed Bandits}
There are also some bandit algorithms~\cite{JoulaniGS2013OLDelayFeedback,ernadeCP2017StochasticDelay,Pike-Burke0SG2018DelayedAggregatedAnonymousFeedback} that investigate the delayed feedback modeling. The objective of these bandit algorithm is to make sequential decisions in order to minimize the cumulative regret. \citet{JoulaniGS2013OLDelayFeedback} analyze the effect of delay on the regret of online learning algorithms and shows that the delay increases the regret in a multiplicative way in adversarial problems, and in an additive way in stochastic problems. \citet{Pike-Burke0SG2018DelayedAggregatedAnonymousFeedback} study the problems of bandits with delayed, aggregated anonymous feedback. \citet{ernadeCP2017StochasticDelay} investigate the stochastic delayed bandit setting and provided a complete analysis under the assumption that the distribution of the delay is known.

\begin{figure}[!t]
    \centering
    \includegraphics[width=\columnwidth]{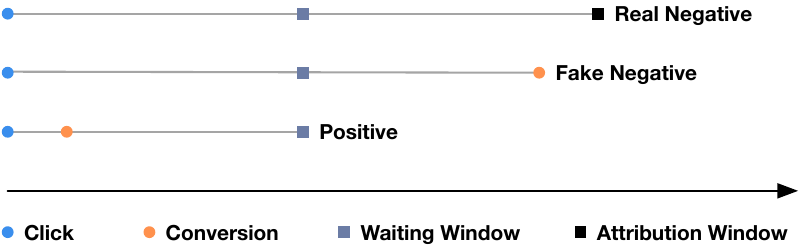}
    \caption{Illustration of the real negative, fake negative and positive. Here, the waiting window is referred to as the time interval between the click time and the time samples ingested into the training pipeline.
    A real negative is the sample that the conversion does not occur. A fake negative is the sample the conversion occurs outside the waiting window but within the attribution window. A positive is the sample that the conversion occurs within the waiting window.}
    \label{fig:sample}
\end{figure}
\section{Preliminary}
In this section, we first give a brief introduction about the background of online CVR modeling. Then we introduce how previous methods address the delay feedback issue for CVR prediction.
\subsection{Background}
In CVR prediction, the model takes input as $(\x, y) \sim (X, Y) $, where $\x$ is the feature and $y \in \{0, 1\}$ is the conversion label. The goal of CVR prediction is to learn a function $f$ with parameter $\theta$ that minimize the the generalization error on sample drawn from this data distribution:
\begin{equation}
    \mathbb{E}_{(\x, y)\sim (X,Y)} [\ell (\x, y; f_{\theta}(\x))],
\end{equation}
where $\ell$ is the loss function.

As the conversion may delay, some of the samples are incorrectly labeled as negatives by traditional supervised learning methods. Denote the duration between the conversion and click as $z$. If the sample will not have conversion eventually, $z = +\infty$. Let $w_1, w_2$ be the length of the waiting window and attribution window, respectively ($w_2>w_1$).
As shown in Figure~\ref{fig:sample}, there exist three types of samples in delayed feedback modeling:
\begin{itemize}
    \item \textbf{Real negatives} ($z>w_2$). Real negatives are samples that the conversion does not occur.
    \item \textbf{Fake negatives} ($w_1<z<w_2$). Fake negatives are samples that the conversion does not occur at the training timing and are incorrectly labeled as negatives due to the waiting window being too short.
    \item \textbf{Positives} ($z<w_1$). Positives have the conversion action inside the waiting window.
\end{itemize}
Note that fake negatives are also positive samples but the conversion don't occur inside the waiting window. A naive strategy to reduce the number of fake negatives is to wait for a sufficiently long window to ensure most of the labels are correct. However, in display advertising systems, the data distribution is dynamically changing, e.g., new ad campaigns are added to the systems.
To capture the change in data, CVR models should update continuously. 
As such, it needs to balance between model freshness and label certainty. Previous methods address this problem by setting a short waiting window to wait for the immediate conversions. To correct the label noise, fake negatives are duplicated with a positive label as soon as the engagement takes place~\cite{KtenaTTMDHYS2019OnlineDFM}. The duplicated samples are ingested into the training pipeline. Hence as shown in Figure~\ref{fig:data_pipeline} (a), the observed distribution contains four parts: real negatives, positives, fake negatives, and duplicated samples that are the duplication of fake negatives but with positive labels. Since directly training with the fake negatives may mislead the direction of optimization, previous methods develop various methods to correct the distribution shift.

\subsection{Fake Negative Weighted}
Due to the disparity between the observed distribution and actual distribution, importance sampling is adopted to correct the distribution shift. \citet{KtenaTTMDHYS2019OnlineDFM} proposes fake negative weighted (FNW) approach.

The loss function of CVR prediction can be written as:
\begin{equation}
\begin{aligned}
    L &= \mathbb{E}_{(\x, y)\sim p} \ell (x, y; f_{\theta}(\x)) \\
    & = \iint p(\x) q(y|\x) \frac{p(y|\x)}{q(y|\x)} \ell (x, y; f_{\theta}(\x)) dx dy \\
    & \approx \iint  q(\x) q(y|\x)  \frac{p(y|\x)}{q(y|\x)} \ell (x, y; f_{\theta}(\x)) dx  dy \\
    & = \mathbb{E}_{(\x, y)\sim q} \frac{p(y|\x)}{q(y|\x)}\ell (x, y; f_{\theta}(\x)),
    \label{eq:importance_sampling}
\end{aligned}
\end{equation}
where $q$ is the biased observed distribution. Due to the ingestion of duplicated positive samples, the observed feature distribution $q(\x)$ is not equivalent to the actual feature distribution $p(\x)$.
FNW assumes that $p(\x) \approx q(\x)$, which may introduce additional bias to the delayed feedback modeling.

In FNW, no waiting window is set and samples are immediately ingested with negative labels. As soon as the user engages with the ad, the same data point will be presented to the model with a positive label.
Hence we can get: $q(\x|y=0) = p(\x)$, $q(\x|y=1) = p(\x|y=1)$, and $q(y=0) = \frac{1}{1+p(y=1)}$. 
The probability of observing a conversion in the biased distribution is:
\begin{equation}
\begin{aligned}
q(y=1|\x) &=  \frac{q(y=1)q(\x|y=1)}{q(y=1)q(\x|y=1) + q(y=0)q(\x|y=0)} \\
          &= \frac{\frac{p(y=1)}{1+p(y=1)} p(\x|y=1)}{\frac{p(y=1)}{1+p(y=1)} p(\x|y=1) + \frac{1}{1+p(y=1)}p(\x)} \\
          & = \frac{p(y=1) p(\x|y=1)}{p(y=1) p(\x|y=1) + p(\x)}\\
          &=  \frac{p(y=1|\x)}{1+p(y=1|\x)}.
    \label{eq:fnw_biased_conversion}
\end{aligned}
\end{equation}
Similarly, the probability of observing a negative sample in the biased distribution can be computed as:
\begin{equation}
    q(y=0|\x) = \frac{1}{1+p(y=1|\x)}.
    \label{eq:fnw_biased_negative}
\end{equation}
By replacing Equation~\ref{eq:fnw_biased_conversion} and Equation~\ref{eq:fnw_biased_negative} in Equation~\ref{eq:importance_sampling}, the loss function can be rewritten as:
\begin{equation}
    \begin{aligned}
    L &= -\sum_{\x, y} y(1+p(y=1|\x))\log f_{\theta}(\x) \\ &+(1-y)p(y=0|\x)((1+p(y=1|\x)) \log (1-f_{\theta}(\x))
\end{aligned}
\end{equation}
Since $p(y=1|\x)$ cannot be accessed, FNW replaces it with the model estimate $f_{\theta}$ and stops the gradient propagation through it during training.

\subsection{Fake Negative Calibration}
Fake negative calibration (FNC) learns a function that directly estimates the biased distribution $q$. 
Then FNC obtains the function $p(y = 1|\x)$ by the following calibration:
\begin{equation}
    p(y = 1|\x) = \frac{q(y = 1|\x)}{1 - q(y = 1|\x)}
\end{equation}


\section{Delayed Feedback Modeling with Real Negatives}
In this section, we present the proposed method, DElayed FEedback modeling with Real negatives ({\defer}) in detail.
We first introduce the designed data pipeline in {\defer}, which duplicates real negatives to correct the distribution shift. Based on the designed data pipeline, {\defer} use importance sampling to correct the distribution shift. Other methods can also be trained with the real negatives and we derive the corresponding loss functions. Finally, we share our deployment experiences for commercial platforms with different conversion attribution mechanisms and production budgets.

\subsection{Data Stream}
As mentioned above, previous continuous training methods ingest fake negative samples to catch the freshness at the expense of noisy signals. To introduce more certainty conversion information, duplicated samples are ingested and importance sampling is developed to correct the distribution shift. 
However, these methods have some issues. First, the duplicated samples change the feature distribution. The approximation $q(\x) \approx p(\x)$ may introduce additional bias. Moreover,  all certain information of the conversion action comes from the positive labels, which are scarce in CVR prediction. This makes the model hard to learn and perform well. To reduce the bias without hurting the freshness, we ingest both the duplicated positive samples and duplicated real negatives to the training pipeline. 

Concretely, in the redesigned data pipeline, we let each sample have a waiting window $w_1$ to wait for the conversion signal. The waiting window can be the same for all samples or sample-specific.
The reason to use a sample-specific window is that different products have different delay distributions. For example, expensive products have longer expected conversion times than cheap products, thus need a longer window to wait for the conversion signal. One strategy to set the sample-specific window is to define multiple window length and train a multi-class classification model $p_{\textrm{length}}(w_1|x)$ and predict the waiting window length as:
\begin{equation}
    w_1 = \arg\max_w p_{\rm{waiting}}(w|\x).
\end{equation}
Samples that have conversion in this window are labeled as positive samples and samples that do not have conversion are fake negatives or real negatives.
For those fake negatives, if the conversion occurs later in the attribution window, we will ingest this sample with a positive label into the pipeline. For those real negatives that the conversion does not occur finally, we will also duplicate and ingest them into the training pipeline. The duplication brings more certainty information of the conversion signal. Thus, there are four kinds of samples in the proposed method, as shown in Figure~\ref{fig:data_pipeline} (b), the real negatives, positives, fake negatives, and duplicated samples with the real label (can either be positive or negative).

\subsection{Loss Function}
Due to the disparity between the observed distribution and the actual distribution, the proposed method uses importance sampling to correct the distribution shift. With the ingestion of duplicated samples, we can get $q(\x) = p(\x)$ and the joint probability $q(\x, y=0)$ can be computed as:
\begin{equation}
\begin{aligned}
q(\x, y=0) &= p(\x, y=0) + \frac{1}{2} p(\x, y=1, z > w_1).
\end{aligned}
\end{equation}
The conditional probability $q(y = 0|\x)$ can be written as:
\begin{equation}
\begin{aligned}
q(y = 0|\x) &= \frac{q(\x, y=0)}{q(\x)} \\
& = \frac{p(\x, y=0) + \frac{1}{2} p(\x, y=1, z > w_1)}{p(\x)} \\
& = p(y=0|\x) + \frac{1}{2} p_{dp}(\x)
\end{aligned}
\end{equation}
where $p_{dp}(\x) = p(\x, y=1, z > w_1|\x)$ is the probability of $\x$ being a fake negative.
Similarly,  $q(y = 1|\x)$ can be written as:
\begin{equation}
    q(y = 1|\x) = p(y=1|\x) - \frac{1}{2} p_{dp}(\x).
\end{equation}

Then according to Equation~\ref{eq:importance_sampling}, we can compute the importance sampling weights for $\x$ as:
\begin{equation}
\begin{aligned}
\frac{p(y = 0|\x)}{q(y = 0|\x)}&=\frac{p(y=0|\x)}{p(y=0|\x) + \frac{1}{2}p_{dp}(\x)} \\
\frac{p(y = 1|\x)}{q(y = 1|\x)}&= \frac{p(y=1|\x)}{p(y=1|\x) - \frac{1}{2}p_{dp}(\x)}.
\end{aligned}
\end{equation}
Thus, the importance weighted CVR loss function can be formulated as:
\begin{equation}
\begin{aligned}
L & = - \sum_{\x, y} y \frac{p(y=1|\x)}{p(y=1|\x) -\frac{1}{2} p_{dp}(\x)} \log f_{\theta}(\x)\\
& +(1-y)\frac{p(y=0|\x)}{p(y=0|\x) + \frac{1}{2}p_{dp}(\x)} \log (1-f_{\theta}(\x)).
\label{eq:fianl_loss}
\end{aligned}
\end{equation}

Since $p(y=1|\x), p(y=0|\x)$ cannot be accessed, we replace $p(y=1|\x), p(y=0|\x)$ with the model estimate $f_{\theta}(x)$ and $1-f_{\theta}(x),$ respectively. For $p_{dp}(\x)$, we train a classifier $f_{dp}$ to predict the probability of $\x$ being a fake negative. As in~\cite{KtenaTTMDHYS2019OnlineDFM}, we stop the gradient propagation through the importance weight. Therefore, Equation~\ref{eq:fianl_loss} can be rewritten as:
\begin{equation}
\begin{aligned}
L & = - \sum_{\x, y} y [\frac{f_{\theta}(\x)}{f_{\theta}(\x) - \frac{1}{2}f_{dp}(\x)}] \log f_{\theta}(\x)\\
& +(1-y)[\frac{1-f_{\theta}(\x)}{1-f_{\theta}(\x) +\frac{1}{2} f_{dp}(\x)}] \log (1-f_{\theta}(\x)),
\label{eq:fianl_loss}
\end{aligned}
\end{equation}
where $[\cdot]$ means the stop gradient operation.

\subsection{Variants}
For other methods, like FNW and FNC, the duplicated real negatives can also be ingested to improve learning. We derive the corresponding loss function for FNW and FNC.

\subsubsection{FNW}
For FNW, we have
\begin{equation}
\begin{aligned}
    q(y = 1|\x) &= \frac{p(y = 1|\x)}{1 + p(y = 1|\x) + p(y = 0|\x)} \\
    & = \frac{p(y = 1|\x)}{2}
\end{aligned}
\end{equation}
\begin{equation}
\begin{aligned}
    q(y = 0|\x) & = \frac{1+p(y = 0|\x)}{1 + p(y = 1|\x) + p(y = 0|\x)} \\
    & = \frac{1 + p(y = 0|\x)}{2}
\end{aligned}
\end{equation}
Then we have
\begin{equation}
\begin{aligned}
\frac{p(y = 1|\x)}{q(y = 1|\x)} &=2 \\
\frac{p(y = 0|\x)}{q(y = 0|\x)} &=\frac{2p(y = 0|\x)}{1 + p(y = 0|\x)}
\end{aligned}
\end{equation}
Finally, the loss function is reformulated as
\begin{equation}
\begin{aligned}
L & = - \sum_{\x, y} 2y\log f_{\theta}(\x)\\
& +(1-y)p(y=0|\x)(\frac{2p(y = 0|\x)}{1 + p(y = 0|\x)}) \log (1-f_{\theta}(\x))
\end{aligned}
\end{equation}

\subsubsection{FNC}
Similarly, for FNC, we have
\begin{equation}
    p(y = 1|\x) = 2q(y = 1|\x)
\end{equation}

\subsection{Production Settings of Real Negatives}
In display advertising, different business scenarios often set different attribution window according to the distinct user behaviors. 
For example, in Alibaba, some business scenarios set 1 day as the attribution window while some business scenarios set 7 days as the attribution window.
It is notable that the delay time of real negatives is equal to the maximum delay time of positive samples and
the model freshness is unlikely to be affected hugely by the real negatives but brings more label certainty to the training algorithms. 
For business scenarios with a long conversion window, the ingestion of duplicated samples may increase additional infrastructure cost of maintaining data caches. As such, we approximate the real negatives by setting a shorter window $w_3$ ($w_1 < w_3 < w_2$) for scenarios with long conversion window. 
Samples that do not have conversion before $w_3$ are approximated as real negatives and ingested into the training algorithm. For business scenarios with short attribution window, we employ no approximation and ingest the real negatives.
In other words, we divide the online delayed feedback modeling into 2 settings, short attribution and long attribution window. For the first one, we use the model above directly. For the second one, we let the real negative samples be ingested after a time window $w_3$ ($w_1 < w_3 < w_2$). Similarly, $w_3$ can be fixed for all samples or predicted by a multi-class classification model:
\begin{equation}
    w_3 = \arg\max_w p_{\rm{attribution}}(w|\x).
\end{equation}

\begin{figure}[!t]
    \centering
    \includegraphics[width=.8\columnwidth]{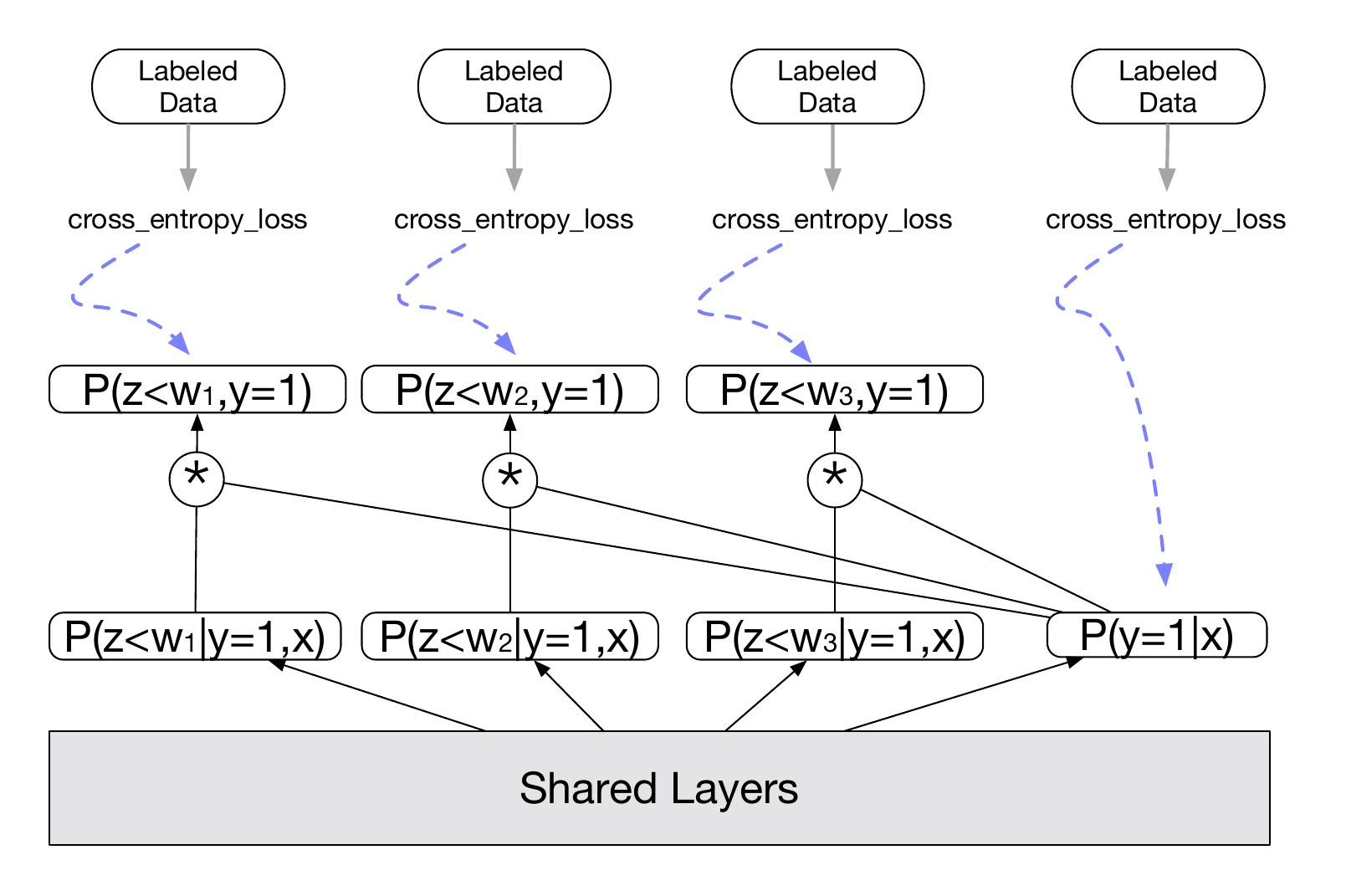}
    \caption{The proposed offline model which uses multi-task learning to improves generalization by leveraging the information contained in the conversion time. One of the head predicts $p(y=1|\x)$, the probability of whether the click will eventually convert. Others predict whether the conversion will occurs within the predefined time windows $w_1, w_2, w_3$.}
    \label{fig:offline_model}
\end{figure}

\subsection{Offline Training Method}
For commercial platforms that apply offline training, we also propose an approach that has been proved to be useful in some scenarios of our display advertising system.

As shown in Figure~\ref{fig:offline_model}, the proposed offline approach uses multi-task learning to improves generalization by leveraging the information contained in the conversion time. Concretely, the model has $N+1$ heads on top of the shared bottom layers. One of the head predicts the probability of whether the click will eventually convert, denoted as $p(y=1|\x)$. Others predict whether the conversion will occurs within the predefined time windows $w_1, w_2, \dots, w_N$:
\begin{equation}
p(z<w_n,y=1|\x)=p(z<w_n|y=1,\x)*p(y=1|\x).
\end{equation}
Denote $y_n$ the label that whether the sample has conversion in $w_n$, the loss function can be written as:
\begin{equation}
\begin{aligned}
L & = - \sum_{n}\sum_{\x} y_n\log p(z<w_n,y=1|\x) \\
&- \sum_{n}\sum_{\x} (1-y_n) \log (1-p(z<w_n,y=1|\x)) \\
&- \sum_{\x} y\log p(y=1|\x) - \sum_{\x} (1-y) \log (1-p(y=1|\x)).
\end{aligned}
\end{equation}
Since samples that near the end of training date may not have all $N+1$ labels, we only update the corresponding parameters according to the observed labels. For a sample, if the predefined $w_n, \dots, w_N$ windows have not reached, we only update the parameters through $p(y=1,z<w_1|\x), \dots, p(y=1,z<w_{n-1}|\x)$ using observed labels while block the parameters of other heads. Note that $p(y=1|\x)$ will also be updated through the gradient of $p(y=1,z<w_1|\x), \dots, p(y=1,z<w_{n-1}|\x)$.
For example, assuming the attribution window as 7 days, we set 1 day, 3 days, 5 days as three time windows. If there is a sample clicked at the 4th day from the last, we only update the parameters from $p(z<w_1,y=1|\x),p(z<w_2,y=1|\x)$. For samples clicked before the 7th day from the last, all the parameters are updated simultaneously.

\section{Experiments}
We evaluate the efficacy of {\defer} in this section. We begin by introducing the setup including the used industrial datasets, construction of data stream, evaluation metrics and compared methods in Sec.~\ref{sec:setting}.  The results and discussion are elaborated in Sec.~\ref{sec:res}. We also investigate the effect of window size in Sec.~\ref{sec:ablation}. Experimental results on production environment and our experience in deploying the model are shown in Sec.~\ref{sec:industrial}.
 \begin{table}[t]
 	\caption{Statistics of Criteo and Taobao-30days dataset.}
 	\label{tab:statistic}
 	\resizebox{\columnwidth}{!}{{\begin{tabular}{c|c|c|c|c|c|c|c}
 	\toprule
 	  Datasets & \#Users  & \#Items & \#Features & \#Conversions & \#Samples & average CVR & Duration \\
 	\midrule
  	Criteo         & - &5443&17&3.6 million&1.59 million&0.2269&60 days\\
 	Taobao-30days & 2 million & 6.5 million & 11 &  12 million    & 116 million &0.1034 & 30 days\\
 	\bottomrule
 	\end{tabular}}}
 \end{table}

 \begin{table*}[t]
 	\caption{Performance comparisons of different models on Criteo and Taobao-30days dataset. Oracle is the upper bound on the best achievable performance as there are no delayed conversions. Bold indicates top-performing method.}
 	\label{tab:exp_result}
 	\resizebox{\textwidth}{!}{{\begin{tabular}{c|c|c|c|c|c|c|c|c|c|c|c|c}
 	
 	\toprule
    \multicolumn{1}{c|}{Datasets} & \multicolumn{6}{c|}{Criteo} & \multicolumn{6}{c}{Taobao-30days}\cr
    \midrule

 	 Metrics &    AUC  & PRAUC & NLL &    RI-AUC  & RI-PR-AUC & RI-NLL &    AUC  & PRAUC & NLL &    RI-AUC  & RI-PR-AUC & RI-NLL \\
 	\midrule
Pre-trained     & 0.8075          & 0.5811          & 0.5425          & 0.00\%           & 0.00\%           & 0.00\%           & 0.6087          & 0.6142          & 0.7693         & 0.00\%           & 0.00\%           & 0.00\%           \\
 Vanilla-NoDup             & 0.7989          & 0.5794          & 0.6149          & -24.29\%         & -2.70\%          & -47.20\%         & 0.6395          & 0.6442          & 0.7754         & 68.44\%          & 75.38\%          & -4.85\%          \\
Vanilla-NoWin & 0.8347          & 0.6189          & 0.4302          & 76.84\%          & 60.00\%          & 73.21\%          & 0.6368          & 0.6395          & 0.7444         & 62.44\%          & 63.57\%          & 19.81\%          \\
Vanilla-Win  & 0.8348          & 0.6251          & 0.4079          & 77.12\%          & 69.84\%          & 87.74\%          & 0.6427          & 0.6474          & 0.6756         & 75.56\%          & 83.42\%          & 74.54\%          \\
Vanilla-RN      & 0.8351          & 0.6266          & 0.4562          & 77.97\%          & 72.22\%          & 56.26\%          & 0.6446          & 0.6496          & 0.6707         & 79.78\%          & 88.94\%          & 78.44\%          \\
FNC             & 0.8347          & 0.6189          & 0.4659          & 76.84\%          & 60.00\%          & 49.93\%          & 0.6368          & 0.6395          & 0.7196         & 62.44\%          & 63.57\%          & 39.54\%          \\
FNC-RN          & 0.8370  & 0.6299 & 0.4103 & 83.33\%  & 77.46\%  & 86.18\%  & 0.6411 & 0.6447 & 0.6710  & 72.00\%  & 76.63\%  & 78.20\%  \\
FNW             & 0.8348 & 0.6262 & 0.4006 & 77.12\%  & 71.59\%  & 92.50\%  & 0.6440  & 0.6477 & 0.6589 & 78.44\%  & 84.17\%  & 87.83\%  \\
FNW-RN          & 0.8372 & 0.6326 & 0.3970  & 83.90\%  & 81.75\%  & 94.85\%  & 0.6458 & \textbf{0.6499} & 0.6592 & 82.44\%  & \textbf{89.70}\%  & 87.59\%  \\

ES-DFM          & 0.8373          & 0.6347          & 0.3956          & 84.18\%          & 85.08\%          & 95.76\%          & 0.6453          & 0.6476          & 0.6560          & 81.33\%          & 83.92\%          & 90.14\%          \\
DEFER           & \textbf{0.8394} & \textbf{0.6367} & \textbf{0.3943} & \textbf{90.11\%} & \textbf{88.25\%} & \textbf{96.61\%} & \textbf{0.6483} & 0.6497 & \textbf{0.6550} & \textbf{88.00\%} & 89.20\% & \textbf{90.93\%} \\
\midrule
Oracle          & 0.8429          & 0.6441          & 0.3891          & 100.00\%         & 100.00\%         & 100.00\%         & 0.6537          & 0.6540           & 0.6436         & 100.00\%         & 100.00\%         & 100.00\%          \\

 	\bottomrule
 	\end{tabular}}}
 \end{table*}

\subsection{Setup}~\label{sec:setting}

\textbf{Datasets}  We experiment with two industrial datasets for evaluation. The characteristics of these datasets are shown in Tabel~\ref{tab:statistic}.
\begin{itemize}
    \item \textbf{Criteo} dataset~\footnote{\url{https://labs.criteo.com/2013/12/conversion-logs-dataset/}} is used for our experiments. Criteo dataset contains more than 15 million  Criteo live traffic data for 60 days. Most of the features are categorical and the continuous features have been made categorical through appropriate quantization. It also includes the timestamps of the clicks and those of the conversions.
    \item \textbf{Taobao-30days} dataset. The real traffic logs for 30 days are collected from the display advertising system of Alibaba. We sub-sample 20\% of the negatives and 1\% of the users. The sampled dataset contains about 120 million samples. For this problem, the conversion is attributed to the last click in a period of 7 days. The dataset has 6 groups of user features and 5 groups of item features.
\end{itemize}

\textbf{Data Stream}
We divide all datasets into two parts to simulate the continuous training paradigm.
The first part of data is used for pre-training the model to obtain a good initialization for continuous training. Note that in case of label leaking, we set the labels as 0 if the conversion occurs in the second part of the data.
The models are then trained and evaluated on the second part.
According to the click timestamp in the datasets, we divide the second part of data into multiple pieces, each of which contains data of one hour. Following the online training manner of industrial systems, we train models on the t-th hour data and test them on the t + 1-th hour. The reported metrics are the weighted average across different hours on streaming data. 

\textbf{Evaluation Metrics}
AUC, PR-AUC and negative log likelihood (NLL) are adopted for evaluation on offline dataset. 
\begin{itemize}
    \item AUC~\cite{Fawcett2006ROC} is a widely used metric in CTR and CVR prediction task~\cite{zhou2018din,zhou2019dien,KtenaTTMDHYS2019OnlineDFM,MaZHWHZG2018ESMM,ShengZZDDLYLZZ2021STAR}, which denotes the probability that a positive sample is ranked higher than a negative sample. 
    \item Area under the Precision-Recall Curve (PR-AUC) is shown to be more sensitive than AUC in the case where there are a lot more negative samples than the positive samples. 
    \item Negative log likelihood (NLL) is a standard measure of a probabilistic model’s quality~\cite{HastieFT2001ESL}. It is sensitive to the absolute value of the predictions~\cite{Chapelle2014DFM}. 
\end{itemize}
We also compute the relative improvement of each model over the pre-trained model. Denote RI-AUC the relative improvement of AUC, RI-AUC$\rm{_{DEFER}}$ can be computed as:
$$
\text{RI-AUC}{\rm{_{Defer}}=\frac{AUC_{Defer}- AUC_{Pre-trained} }{AUC_{Oracle} - AUC_{Pre-trained}}} \times 100\%,
$$
where ${\rm{AUC_{Oracle} - AUC_{Pre-trained} }}$ means how much we can improve over the pre-trained model and ${\rm{AUC_{Defer}- AUC_{Pre-trained}}}$ means how much {\defer} improves over the pre-trained model. Note that the closer the relative improvement to 100\%, the better the method performs. The relative improvement of PR-AUC (RI-PR-AUC) and NLL (RI-NLL) are computed similarly.

For online A/B testing, we report the CVR. The online CVR reflects the actual performance that determines which approach is the most suitable for the system.

\begin{figure*}[!t]
	\centering
	\begin{minipage}[h]{.33\textwidth}
		\centering
		\includegraphics[width=\textwidth]{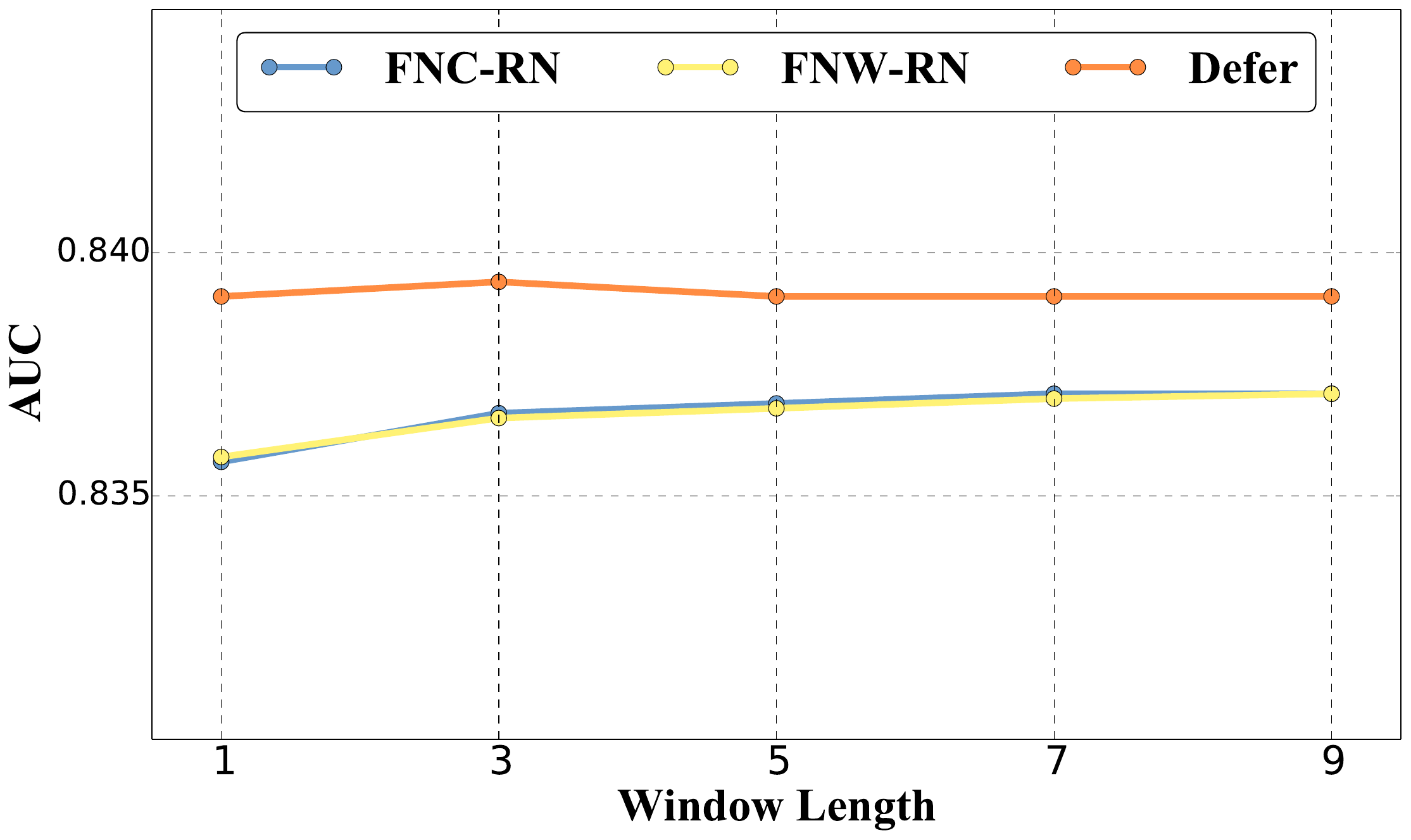}
		\mbox{({\it a}) {AUC}}
	\end{minipage}
    \begin{minipage}[h]{.33\textwidth}
		\centering
		\includegraphics[width=\textwidth]{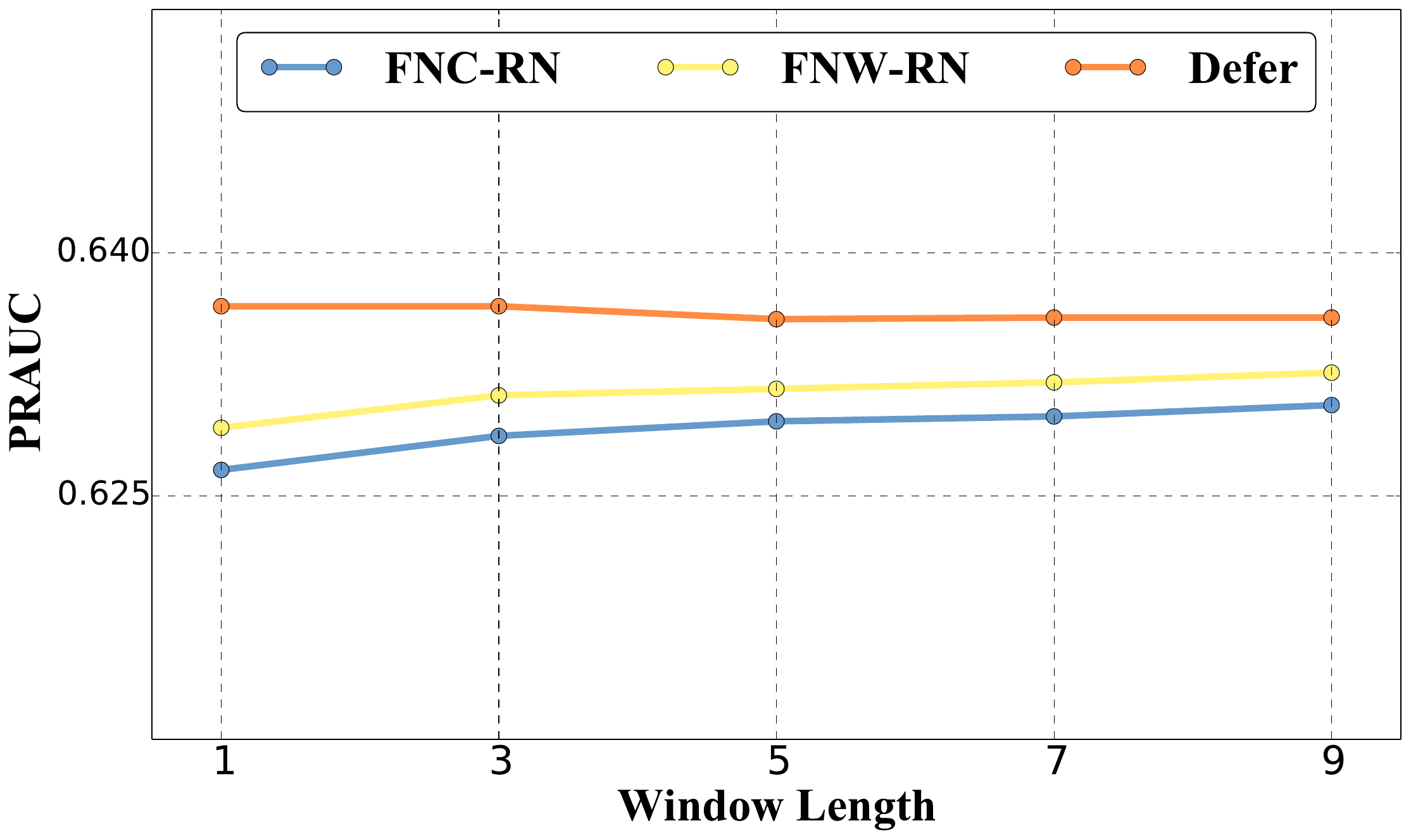}
		\mbox{ ({\it b}) {PR-AUC}}
	\end{minipage}
    \begin{minipage}{.33\textwidth}
    \centering
		\includegraphics[width=\textwidth]{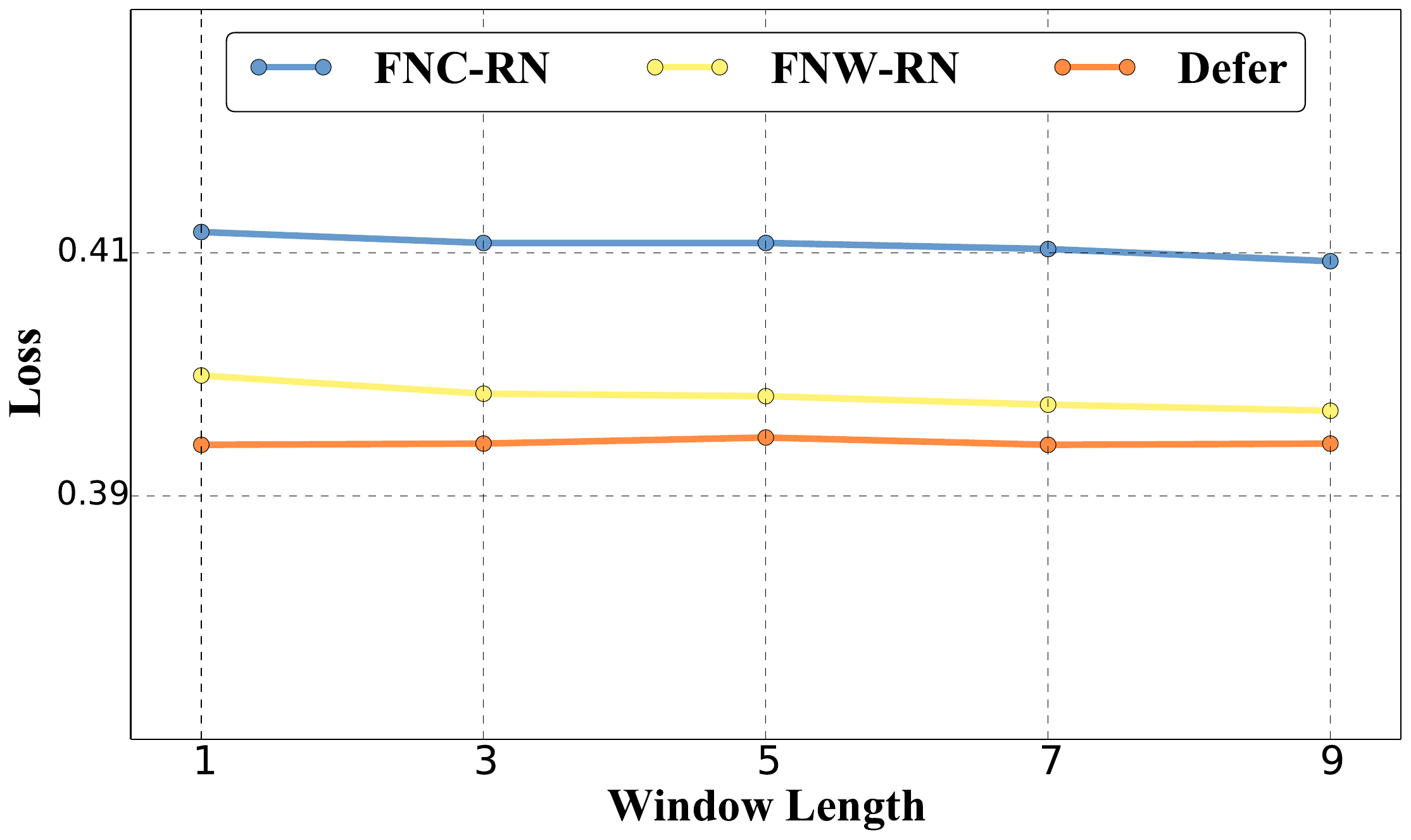}
		\mbox{ ({\it c}) {Loss}}
	\end{minipage}
	\caption{The effect of different window length for approximating the real negatives.}
	\label{fig:eval_diff_window_length}
\end{figure*}

\textbf{Compared Methods}
In order to evaluate the performance of the proposed {\defer} approach, the following methods are implemented and compared:
\begin{itemize}
    \item \textbf{Pre-trained}: The pre-trained model is trained on the data before the streaming data.
    \item \textbf{Vanilla-NoDup}: Vanilla-NoDup is the online learning model trained on streaming data with a waiting window but without duplicated samples. A sample that has conversion in the waiting window is labeled as positive, else as negative.
    \item \textbf{Vanilla-NoWin}: This method is trained on streaming data without waiting window. Samples are immediately ingested with negative labels and then insert a positive duplication when conversion happens. 
    \item \textbf{Vanilla-Win}: Vanilla-Win is trained on streaming data with a waiting window. The positive samples are duplicated and ingested after the window. 
    \item \textbf{Vanilla-RN}: Vanilla-RN is also trained on streaming data with a waiting window. Positives and real negatives are duplicated and ingested after the window.
    
    \item \textbf{FNW}~\cite{KtenaTTMDHYS2019OnlineDFM}: FNW is trained on the same streaming data as Vanilla-NoWin. The loss function is the importance weighted loss.
    \item \textbf{FNW-RN}: Compared with FNW, FNW-RN additionally uses the duplicated real negatives.
    \item \textbf{FNC}~\cite{KtenaTTMDHYS2019OnlineDFM}: FNC is trained on the same streaming data as Vanilla-NoWin.
    \item \textbf{FNC-RN}: Compared with FNC, FNC-RN additionally use the duplicated real negatives.
    \item \textbf{ES-DFM}~\cite{YangLHZZZT2020ESDFM}: ES-DFM is trained on the same streaming data as Vanilla-Win. The loss function is the ES-DFM loss.
    
    \item \textbf{DEFER}: The proposed {\defer} method is trained on the same streaming data as Vanilla-RN but with the proposed loss function.
    \item \textbf{Oracle}: The oracle can access the true label, i.e., it can look into the future to determine if there will be a conversion. This is the upper bound on the best achievable performance as there are no delayed conversions for oracle.
\end{itemize}
Note that all Vanilla-$\{\cdot\}$ methods use the vanilla log loss function while FNW, FNW-RN, ESDFM, and {\defer} use the importance weighted loss function. FNC and FNC-RN are trained with the vanilla log loss function but apply the probability calibration.

\textbf{Implementation Details}
We implement the CVR model as a deep neural network. The features are transformed to embedding vectors firstly and then fed to fully-connected layers. The hidden units are fixed for all models with hidden size \{256, 256, 128\}. The activation functions are Leaky ReLU~\cite{Maas2013ReLU}. For each hidden layer, we also apply batch normalization~\cite{IoffeS2015BN} to accelerate the model training. All  methods are trained with Adam~\cite{KingmaB2014Adam} for optimization. Following ES-DFM~\cite{YangLHZZZT2020ESDFM}, the waiting window $z_1$ is set as 0.25 hour.
To give a fair comparison, all hyper-parameters are tuned carefully for all compared methods. For methods that use the real negatives, the window  $z_3$ to approximate the real negatives is also tuned fairly and we report the results at the best window length.
\subsection{Results}~\label{sec:res}
We evaluate all approaches on the Criteo and Taobao-30days dataset. The result is reported in Table~\ref{tab:exp_result}. Except for the close PRAUC compared with FNW-RN on Taobao-30days, the proposed method almost consistently outperforms other approaches on all performance metrics, which validates the effectiveness of {\defer}. We analyze the results from various aspects. 

\textbf{Continuous training with duplicated samples contributes to the performance lift.} It is notable that all continuous training methods, except for Vanilla-NoDup, outperform the pre-trained method, especially on NLL.  Compared with FNW and FNC, Vanilla-NoDup performs poorly since positives outside the waiting window are not corrected, making the observed distribution deviate much from the actual distribution. These results demonstrate the importance of continuous training on real-time samples and duplicated samples. We also notice that Vanilla-Win is better than Vanilla-NoWin, which suggests \textbf{the waiting window in the training pipeline is crucial} as it provides more label correctness. The comparison between FNW and ES-DFM also indicates the importance of the waiting window.

Note that all methods using the duplicated real negatives (Vanilla-RN, FNC-RN, FNW-RN, {\defer}) achieve better performance than the counterpart methods trained without the real negatives (Vanilla-Win, FNC, FNW, ES-DFM). \textbf{The consistent superiority validates the effectiveness of ingesting real negatives} as it reduces the bias and introduces more certain conversion information. We also notice that the Vanilla-NoDup performs worse on Criteo than on Taobao-30days, which result from the different conversion distributions. On Criteo, the delayed feedback is more serious, where only 35\% conversions happen in 0.25 hours. By contrast, about 55\% conversions take place in 0.25 hours on Taobao-30days. This also validates the importance of duplicating samples with real labels when facing severely delayed feedback.

The comparison between Vanilla-NoWin, FNC, and FNW indicates \textbf{the necessity of designing appropriate loss functions}. The gap between Vanilla-Win and ES-DFM and the gap between Vanilla-RN and {\defer}  also validate the effect of importance sampling to correct the distribution shift. Finally, the relative metrics among Oracle, Pre-trained, and {\defer} demonstrate that the gap is narrowed down by the proposed method significantly. 


\subsection{Analysis of Window Length $z_3$}\label{sec:ablation}
A short window for approximating the real negatives can reduce the cost of maintaining data caches but could also introduce fake negatives. We study the effect and sensitivity of different window length $z_3$ for approximating the real negatives.
 
To this purpose, FNW-RN, FNC-RN, and {\defer} are trained with $z_3$ = \{1, 3, 5, 7, 9\} days on the Criteo dataset. As shown in Figure~\ref{fig:eval_diff_window_length}, 
{\defer} achieves the best performance with window length $z_3 = 3$, while FNW-RN and FNC-RN achieve the best performance with  $z_3 = 9$. Note that all evaluation metrics of the three methods stay stable with window lengths longer than 3 days. The results suggest that we can approximate the real negatives with a relatively short window length. It can also be inferred that the freshness of the training data is already captured when the samples are firstly ingested into the training pipeline. The duplicated samples mainly help eliminate the bias in the observed distribution and reduce the uncertainty. Since the ingestion of real negatives with long window length improve little but consume much more resources, we can choose a suitable short window in industrial practice.

\subsection{Deployment}\label{sec:industrial}

\subsubsection{Online Serving with Continuous Training}

To verify the effectiveness of our continuous training method, we conduct experiments on 2 scenarios of the display advertising system of Alibaba. In both scenarios, the goal of the CVR model is to estimate the probability of \textit{Adding To Cart} and the attribution window of conversion is 1 day. Compared with the pre-trained model, we observe 8.5\% improvement on CVR during the online A/B testing. The model has been successfully deployed into production, serving the main traffics for these scenarios.

\subsubsection{Online Serving with Offline Training}
We evaluate the proposed multi-task approach on a scenario of our display advertising system that uses offline training. The goal of CVR prediction is to estimate the probability of \textit{Purchase} and the attribution window of conversion is 7 days. The model is trained on data of 30 days and test on the data of the following date. The AUC of the proposed approach and the baseline model using vanilla log loss are 0.8385 and 0.8347 respectively. During the online A/B testing, we observe 6\% improvement on CVR. The model has also been successfully deployed into production, serving the main traffic.

\section{Conclusion}
In this paper, we have investigated the problem of online delayed feedback modeling. Compared with previous approaches, the proposed {\defer} ingests duplicated real negatives without changing the original feature distribution. The ingestion of duplicated real negatives provides more label correctness, facilitating the modeling of delayed feedback. With the ingestion of real negatives, {\defer} further employs importance sampling to correct the distribution shift. Apart from the proposed method for online CVR prediction, we also share our deployment experiences for commercial platforms with different conversion attribution mechanisms and budgets. Experiment result on industrial dataset validates the superiority of the proposed methods. Up to now, the proposed methods have been successfully deployed in the display advertising system of Alibaba. The practical experience gained in our deployment generalizes to other setups, thus of interest to both researchers and industrial practitioners.

\section{Acknowledgments}
We appreciate the helpful discussion as well as the support for experiments from Jia-Qi Yang.

\bibliographystyle{ACM-Reference-Format}
\balance
\bibliography{rn_odl}


\end{document}